\documentclass[letterpaper, 10 pt, conference]{ieeeconf}  

\IEEEoverridecommandlockouts                              

\makeatletter
\let\NAT@parse\undefined
\makeatother
\usepackage{amsmath} 
\usepackage{amssymb}  
\usepackage{subcaption, graphicx}
\usepackage{hyperref}
\usepackage{cite}

\title{\LARGE \bf
Photorealistic Robotic Simulation using Unreal Engine 5 for Agricultural Applications
}

\author{Xingjian Li$^{1}$ and Lirong Xiang$^{2,*}$
\thanks{*This work is supported by USDA NIFA Hatch project (7005224).
}
\thanks{$^{1}$Xingjian Li is with the Department of Electrical and Computer Engineering,
        North Carolina State University, USA.
        {\tt\small xli228@ncsu.edu}}%
\thanks{$^{2}$Lirong Xiang is with the Department of Biological and Agricultural Engineering, North Carolina State University,
        USA.
        {\tt\small lxiang3@ncsu.edu}}%
}

\begin{document}

\maketitle
\thispagestyle{empty}
\pagestyle{empty}

\begin{abstract}

This work presents a new robotics simulation environment built upon Unreal Engine 5 (UE5) for agricultural image data generation. The simulation utilizes the state-of-the-art real-time rendering engine to provide realistic plant images which are often used in agricultural applications. This study showcases the rendering accuracy of UE5 in comparison to existing tools and assesses its positional accuracy when integrated with Robot Operating Systems (ROS). The results indicate that UE5 achieves an impressive average distance error of 0.021mm when compared to predetermined setpoints in a multi-robot setup involving two UR10 arms.


\end{abstract}

\section{INTRODUCTION}

Agricultural robots often require extra time management and careful planning for their experiments because many tasks are time-consuming and time-dependent. Therefore, data collection and robot simulations are especially important to avoid these constraints. While there are agricultural robotics simulations for image-based tasks\cite{shafiekhani_vinobot_2018,tsolakis_agros_2019}, their Gazebo-based rendering lacks the detailed rendering provided in modern game engines such as Unity or UE5. This work presents a UE5 and ROS simulation to provide photorealism on top of common robot simulation tools, with an experiment and analysis of a multi-robot setup of two UR10 arms tasked to scan a weed plant.


\section{RELATED WORKS}

UE5 provides state-of-the-art real-time photorealistic rendering for the game and entertainment industry. Compared to UE4, UE5 provides Lumen and Nanite for performant detailed rendering, and high-quality, free assets from Quixel Megascans \cite{noauthor_unreal_nodate}. In recent years, photorealistic game engines such as Unity and UE (particularly UE4) were used for robotics simulation in areas such as autonomous driving \cite{dosovitskiy_carla_2017}, ground\cite{anand_openuav_2021, mania_framework_2019}, aerial\cite{shah_airsim_2017, guerra_flightgoggles_2019}, underwater robots\cite{anand_openuav_2021, potokar_holoocean_2022}, and industrial arms \cite{zuo_craves_2019, erdei_design_2022} for the purposes of simulation, synthetic data generation \cite{zuo_craves_2019, mania_framework_2019, martinez-gonzalez_unrealrox_2020}, and user training via virtual reality \cite{erdei_design_2022}, many of these tasks are met with success unique to the engine choice. Moreover, these game engines have demonstrated successful transfers from synthetic images to real images across numerous tasks. \cite{mishra_task2sim_2022, martinez-gonzalez_unrealrox_2020}.

\section{SIMULATION SETUP}
The simulation is set up as a digital twin with one set in UE5 and the other in ROS. The two communicate using rosbridge and ROSIntegration \cite{github_rosintegration} through WebSockets.
\subsection{UE5 Environment} \label{sec:UE5 Environment}

\begin{figure}[b]
    \centering
    \includegraphics[height=4.5cm]{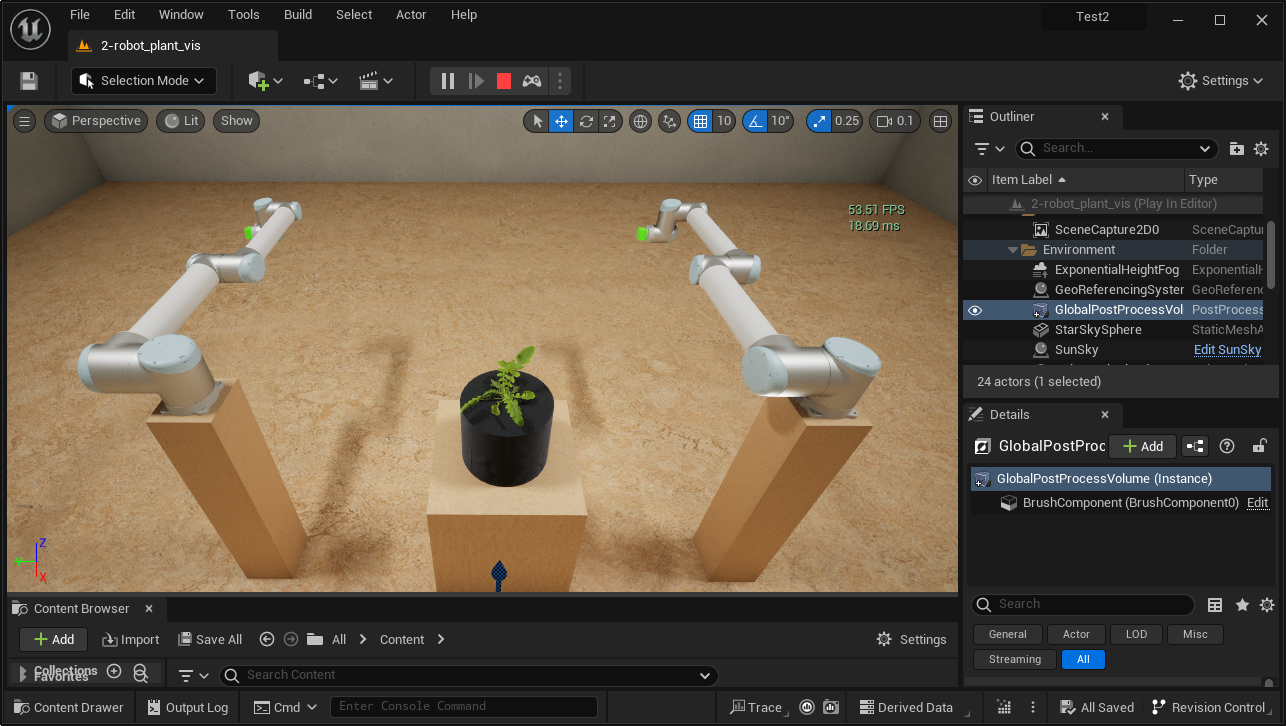}
    \caption{\textbf{Unreal Engine 5 Environment}}
    \label{fig:UE5_Environment}
\end{figure}

The environment, as shown in Fig. \ref{fig:UE5_Environment}, simulates a simple, indoor space with two robot stands and a table in-between for the plant. The plant and materials are downloaded from Quixel Megascans. There are four light sources, two of them are placed as ceiling lights for room lighting, and the other two are placed on the robot stand for plant lighting. Lumen and raytraced shadows are used to provide photorealistic lighting and shadows on the plant for RGB and depth data. For segmentation images, a copy of this room is made with constant base color to capture the flat colors.

\begin{figure*}[ht!]%
\centering
\begin{subfigure}[t]{.5\textwidth}
\centering
\includegraphics[height=2.45cm]{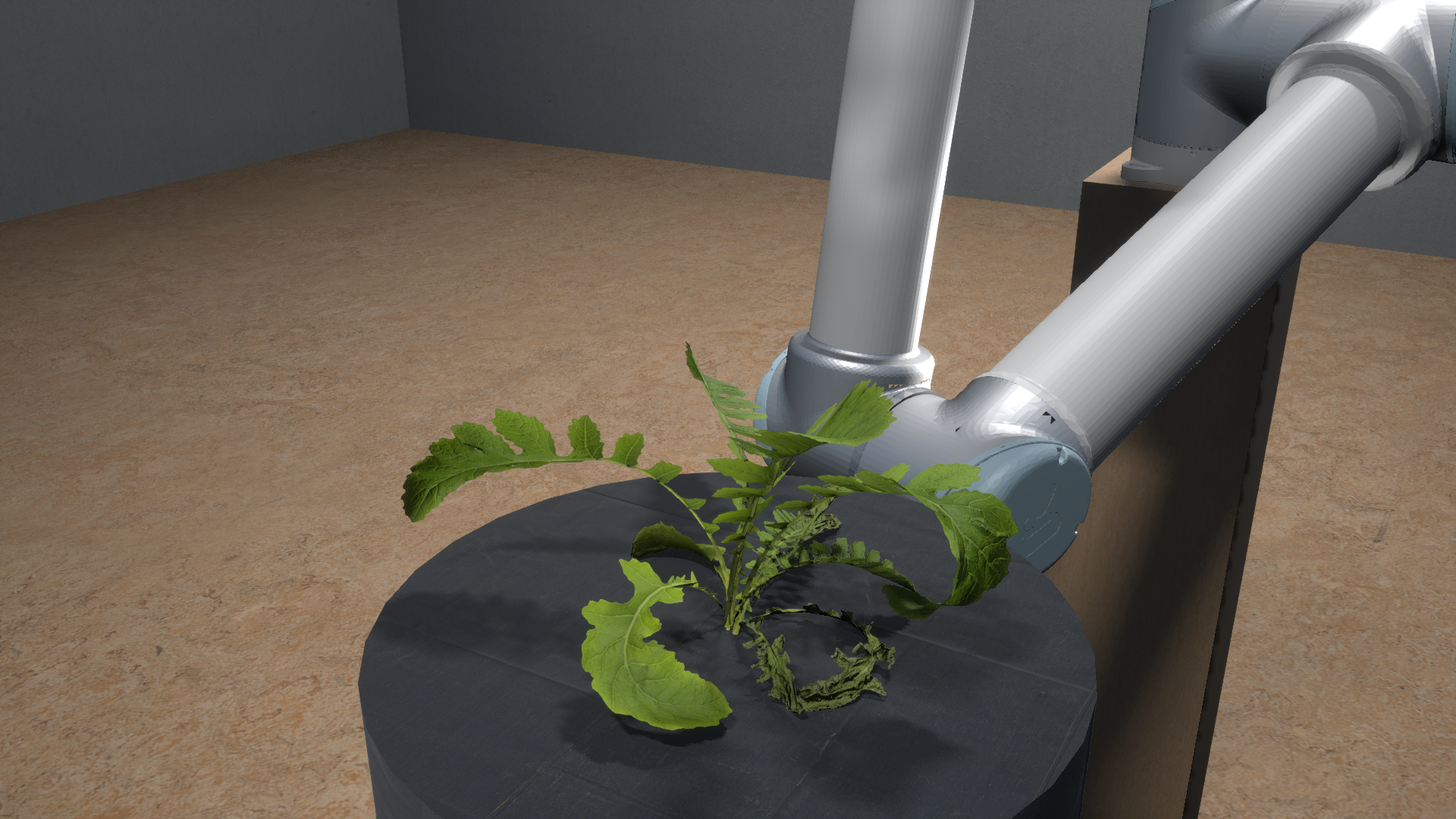}%
\includegraphics[height=2.45cm]{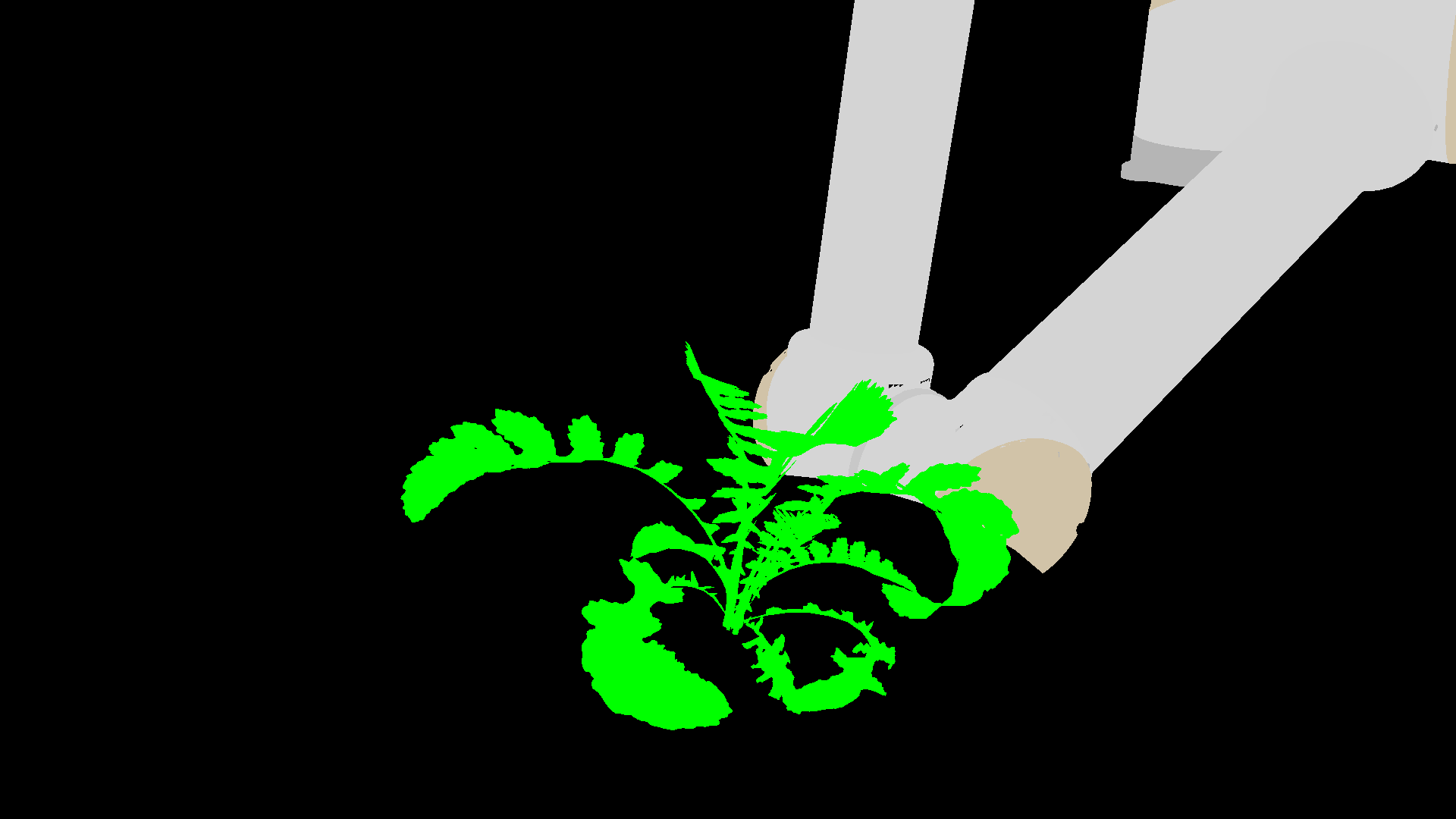}%
\caption{Rendered RGB and segmentation images}
\label{fig:simulation}%
\end{subfigure}%
\begin{subfigure}[t]{.5\textwidth}
\centering
\includegraphics[height=2.45cm]{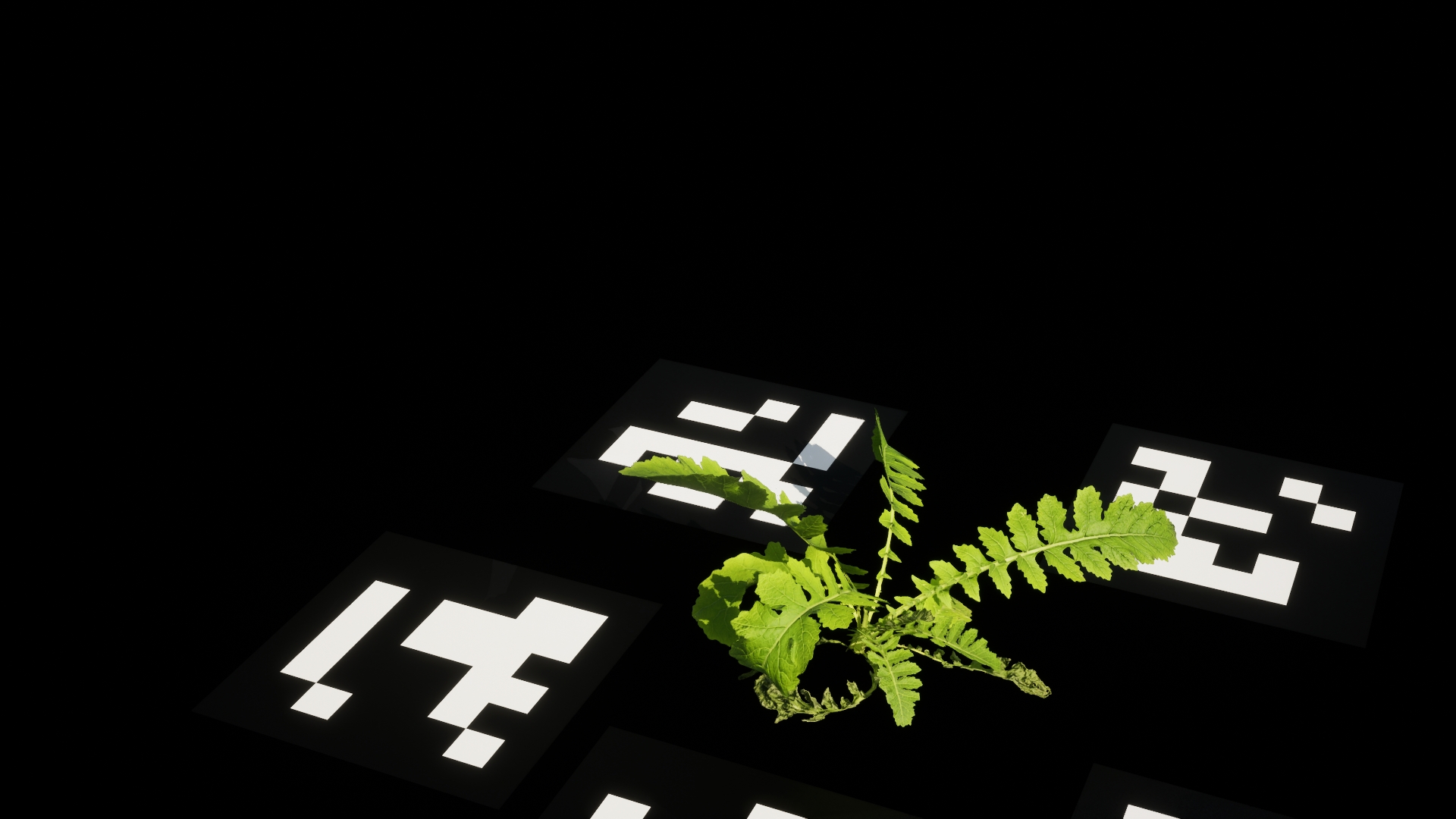}%
\includegraphics[height=2.45cm]{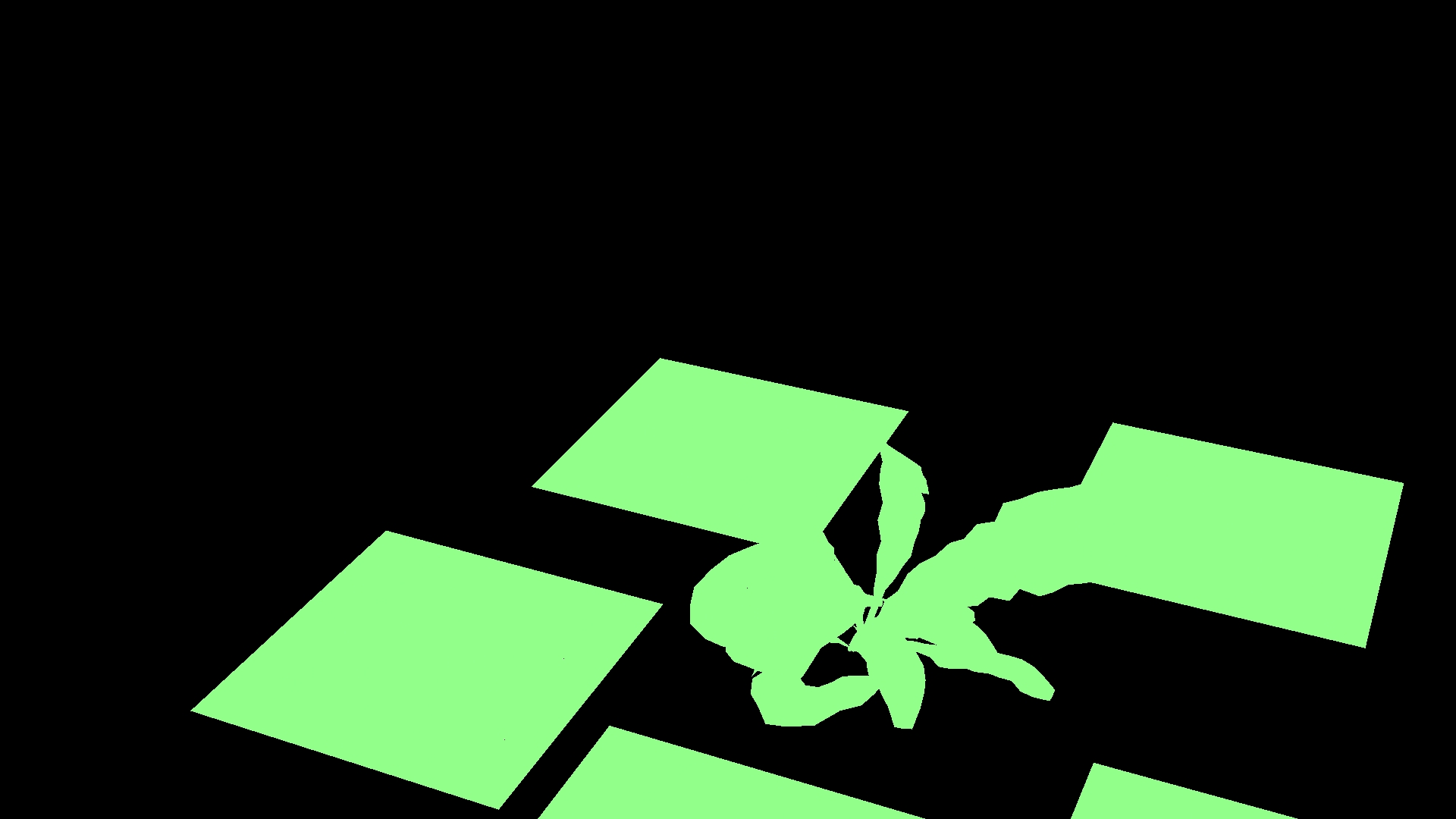}%
\caption{EasySynth plugin of the same plant in different environment}
\label{fig:easysynth}%

\end{subfigure}%
\caption{\textbf{Rendered RGB and segmentation image comparison.} The segmentation colors for a) are determined by the base color of the material of the copied room as described in Section \ref{sec:UE5 Environment}. Notice the lack of details in b)'s segmentation compared to a)'s segmentation, where the plugin removes the transparency of the material by replacing it with new material.}
\label{fig:sample_data}
\end{figure*}
\subsection{UE5 \& ROS Robots}

The UR10 is constructed as a UE5 blueprint based on the URDF files and models in the ur\_description repository\cite{charles_universal_2022}, the DAE mesh is converted to FBX format for importing, and the joints are modeled using PhysicsConstraints components with angular motion. SceneCapture2Ds are attached to the end effector to take RGB, segmentation, and depth images. The robot blueprints can be placed in the scene with different IDs to receive different messages. Coding is done using UE5 visual scripting with ROS messages in C++. The arm physics in UE5 is turned off except for its environmental collisions, due to difficulty in tuning motor strengths and UE's limitation in physics accuracy \cite{spryn_simulating_2019}.


MoveIt \cite{sucan_moveit_nodate} is used for robot planning and visualization. The multi-robot is almost identical to the UE5 setup with the end effector being a simple cylinder. Three boxes are spawned to ROS during runtime to mimic the stand collisions in the UE5 scene in Fig. \ref{fig:UE5_Environment}.

\subsection{Communication}

To replicate movement from ROS to UE5, the /joint\_state topic containing ROS joint angles is processed and republished to each robot arm in UE5. ROS does not verify the state of the UE5 joints, but the world position of the end effector is collected for experiments and analysis. ROS sends a boolean message to UE5 for image data collection, which is saved as 1080p image files under the UE5 project folder as a buffer for processing. 

\section{EXPERIMENT}

We used the simulation to take synthetic images around the plant to generate RGB, segmentation, and depth images along with camera and robot poses. The setpoints, positions, and orientations, were predetermined with spherical and cylindrical motion around the plant. The linear and joint trajectories were generated for the two arms during the simulation. UE5 recorded the end effector world position and the /tf to the ROS end effectors was also recorded at the same time for trajectory comparison. The simulation was ran on a i7-12700k, RTX3060, and 32GB memory desktop.

\begin{figure}[b]
    \centering
    \includegraphics[height=4.5cm]{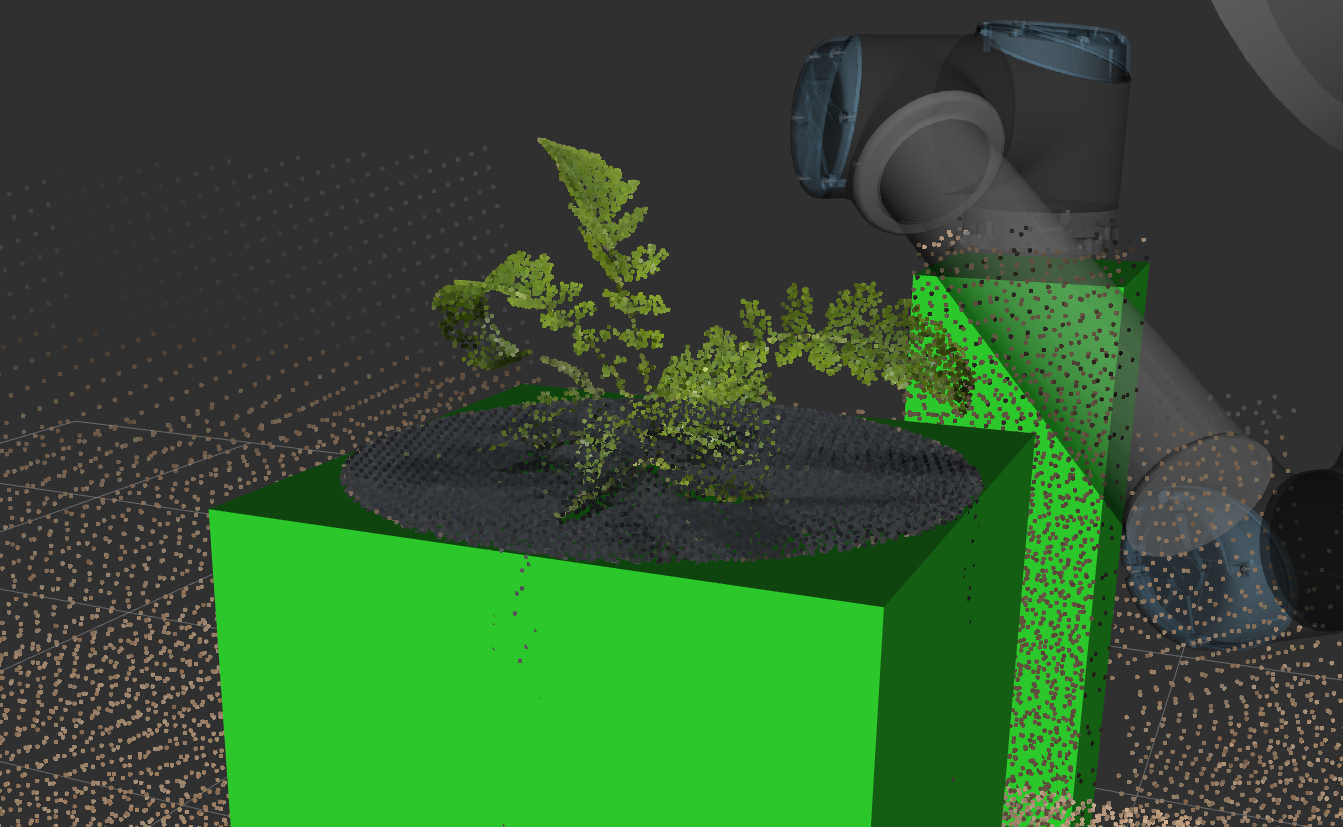}
    \caption{\textbf{Point clouds from UE5 RGBD Images displayed in RViz during the simulation.} The point cloud below the plant is circular because it is a cylinder in UE5.}
    \label{fig:point_clouds}
\end{figure}
\begin{figure}[b]
    \centering
    \includegraphics[width=.49\columnwidth]{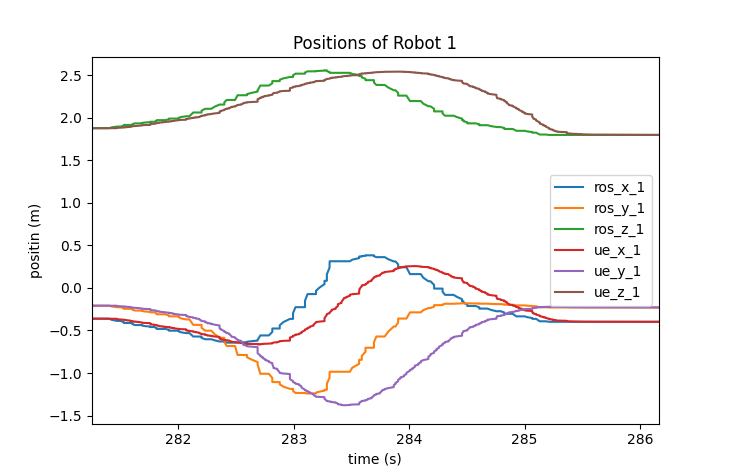}
    \includegraphics[width=.49\columnwidth]{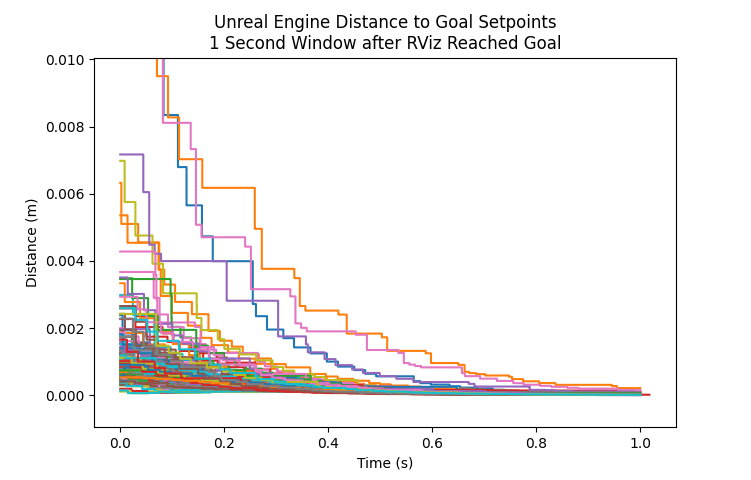}
    \caption{\textbf{End effector trajectories.} Left: world positions of the robot 1 end effector for UE5 and ROS of one trajectory with the highest same-time distance error (0.77m at one point). Right: the 1-second window of UE5 distance error after ROS has reached one of 120 setpoints.}
    \label{fig:accuracy}
\end{figure}
\section{RESULTS}

\subsection{Synthetic Images}
Fig. \ref{fig:simulation} displays the RGB and segmentation images captured by one of the arms at the same location. The raytraced lights and shadows provide the plant with realistic shading for healthy leaves and dried leaves. Particularly in the details of soft and hard shadows from different light sources in the room. The segmentation properly segments individual leaves, while EasySynth \cite{noauthor_easysynth_2023}, a UE5 image data generation plugin, would incorrectly segment the gaps in between the leaves as shown in Fig. \ref{fig:easysynth}. The RGB and depth information were processed into point clouds and visualized in RViz during the simulation as shown in Fig. \ref{fig:point_clouds}. The RGB and segmentation were saved as PNG files while depth was saved as EXR files with scene depth information (true depth in world distance).

\subsection{Simulation Accuracy}

The UE5 simulation has delays but is accurate to ROS trajectories as shown in Fig. \ref{fig:accuracy}. Visually the end effector follows a similar trajectory during the movement, with UE5 lagging behind ROS due to open-loop control from ROS joints to UE5 joints and lack of UE5 motor strength parameters. When ROS reaches one of the 120 setpoints, the UE5 to setpoint L2 distance error is on average 1.769mm with a max of 19.148mm error, and the 1-second error is on average 0.021mm with a max of 0.211mm.

\section{CONCLUSIONS AND FUTURE WORK}

Overall, the simulator shows promising performance in simulating armed robots and plant photorealism that could be beneficial for agricultural robotics development. The simulator allows for the verification of agricultural image processing algorithms with realistic images and without planning for experiment times. Currently, this simulator has not implemented gripper manipulation and wheeled mobility, so harvesting tasks cannot be fully simulated yet. In future work, we plan for these features and also examine the current simulation with a real setup for 3D plant reconstruction to increase the variety of plant assets available for the simulator.


\addtolength{\textheight}{-12cm}   








\bibliographystyle{IEEEtran}
\bibliography{Workshop_UE5_Sim}

\end{document}